\pdfoutput=1

\documentclass[11pt]{article}

\usepackage[final]{acl}

\usepackage{times}
\usepackage{latexsym}

\usepackage[T1]{fontenc}

\usepackage[utf8]{inputenc}

\usepackage{microtype}

\usepackage{inconsolata}

\usepackage{graphicx}

\usepackage{inconsolata}
\usepackage{multirow}
\usepackage{hhline}
\usepackage{pifont}

\usepackage{algorithm}
\usepackage{algpseudocode}
\usepackage{amssymb}
\usepackage{amsmath}
\usepackage{mathtools}

\usepackage{booktabs}
\usepackage{xspace}

\usepackage{xcolor}

\usepackage[cjk]{kotex}
\usepackage{graphicx}
\usepackage[normalem]{ulem}
\useunder{\uline}{\ul}{}

\definecolor{orangebg}{RGB}{255,229,207}
\definecolor{orangetxt}{RGB}{229,90,0}
\definecolor{bluebg}{RGB}{217,227,242}
\definecolor{bluetxt}{RGB}{43,85,157}

\newcommand{\category}[1]{\textcolor{bluetxt}{#1}}
\newcommand{\sentiment}[1]{\textcolor{orangetxt}{#1}}
\newcommand{\aspect}[1]{\colorbox{bluebg}{#1}}
\newcommand{\opinion}[1]{\colorbox{orangebg}{#1}}

\title{Evaluating Span Extraction in Generative Paradigm:\\A Reflection on Aspect-Based Sentiment Analysis}

\author{Soyoung Yang \\
  KAIST AI\\
  \texttt{sy\_yang@kaist.ac.kr} \\\And
  Won Ik Cho\thanks{Corresponding author} \\
  SAIT, Samsung Electronics \\
  \texttt{wonik.cho@samsung.com} \\}

\begin{document}
\maketitle
\begin{abstract}
In the era of rapid evolution of generative language models within the realm of natural language processing, there is an imperative call to revisit and reformulate evaluation methodologies, especially in the domain of aspect-based sentiment analysis (ABSA). This paper addresses the emerging challenges introduced by the generative paradigm, which has moderately blurred traditional boundaries between understanding and generation tasks. 
Building upon prevailing practices in the field, we analyze the advantages and shortcomings associated with the prevalent ABSA evaluation paradigms. Through an in-depth examination, supplemented by illustrative examples, we highlight the intricacies involved in aligning generative outputs with other evaluative metrics, specifically those derived from other tasks, including question answering.
While we steer clear of advocating for a singular and definitive metric, our contribution lies in paving the path for a comprehensive guideline tailored for ABSA evaluations in this generative paradigm. In this position paper, we aim to provide practitioners with profound reflections, offering insights and directions that can aid in navigating this evolving landscape, ensuring evaluations that are both accurate and reflective of generative capabilities.
\end{abstract}

\section{Introduction}

\begin{figure}[t!]
    \centering
    \includegraphics[trim={0cm 0.4cm 0cm 0.2cm}, clip=true, width=.9\columnwidth]{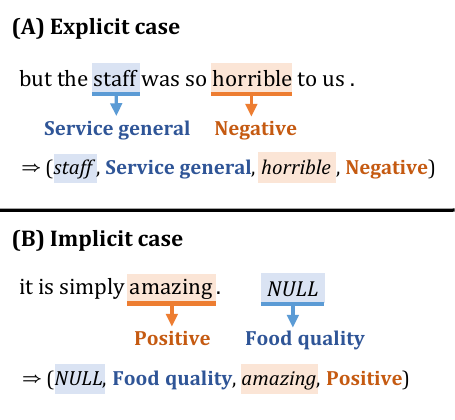}
    \caption{Aspect sentiment quad prediction (ASQP) examples from ACOS~\cite{cai-2021-acos} rest16 dataset. Each quadruple is extracted from a given sentence in the order of (aspect term $a$, \category{aspect category $c$}, opinion term $o$, \sentiment{sentiment polarity $s$}). 
    Example (A) is an explicit case where the mentions of aspect and opinion terms are described in the given sentence, while (B) is an implicit one where the aspect term is not found in the sentence.  
    }
    \label{fig:example}
\end{figure}

Extracting information from user-generated reviews is pivotal and essential in real-world applications. In response to this demand, aspect-based sentiment analysis (ABSA), a task of extracting various aspects of sentiment information from user reviews, has emerged and evolved with various studies~\cite{liu2012sentiment,chebolu2022survey}.
The contemporary iterations of ABSA typically refer to the aspect sentiment quad prediction (ASQP) 
framework~\cite{zhang-2021-asqp-paraphrase}, encompassing four elements: aspect term, aspect category, opinion term, and sentiment polarity\footnote{In this paper, our discussions operate under the default ASQP configuration that predicts all four elements.}.
Fig.~\ref{fig:example} shows the two examples that extract one quadruple for each input sentence. In example (A),
the aspect and opinion terms, i.e., ``\textit{staff}'' and ``\textit{horrible}'', are span factors directly extracted from source sentence(s), whereas aspect category and sentiment polarity, i.e., ``\category{Service general}'' and ``\sentiment{Negative}'', reflect their corresponding categorical attributes. 
Fundamentally, ABSA is rooted in comprehension; to tackle the problem accurately, a model must grasp both context and underlying intent.
However, in the era of generative language models (GLM), researchers encounter a novel realm of challenges in both extracting and classifying these four elements. 

Adding to the complexity, recent ABSA methodologies, such as ACOS dataset~\cite{cai-2021-acos}, have incorporated a ``\textit{NULL}'' annotation to address instances where span tagging is not applicable, echoing the challenges faced with unanswerable scenarios in question-answering (QA) tasks~\cite{rajpurkar-etal-2018-know-squad2}.
As shown in example (B) of Fig.~\ref{fig:example}, the aspect term is tagged as ``\textit{NULL}'' because the term is not represented as an entity explicitly, unlike the opinion term ``\textit{amazing}''.
Moreover, a broader spectrum of domains is covered with the latest expansions in the MEMD-ABSA dataset~\cite{cai-2023-memdabsa}, introducing new domains of books, clothing, and hotel, along with the explored areas of laptop and restaurant. 
This inclusivity presents a diverse range of tendencies across domains, and the potential aspect and opinion candidates similarly exhibit extensive variability.

Despite the ongoing research in models and datasets to perform ABSA tasks, there has been less discussion on the evaluation scheme of their outcomes.
In fact, evaluating the output produced by the ABSA model(s), particularly implemented using the extract-classify-ACOS methodology~\cite{cai-2021-acos}, poses intricate hurdles, primarily because it encompasses two concurrent abilities of extraction and classification, each demanding simultaneous attention.
As shown in Fig.~\ref{fig:example}, the ABSA task involves extraction and classification processes, leading to the formation of pairs: (aspect term, \category{aspect category}) and (opinion term, \sentiment{sentiment polarity}), where the former requires extraction and the latter necessitates categorizing ability. 
On the surface, this dichotomy, pairing the aspect term with the aspect category and the opinion term with the sentiment polarity, seems logically structured. 
However, this division doesn't necessarily indicate that the decision-making process for one attribute is wholly independent of the other since the spans of aspects and opinions terms influence the decision process of each other and sometimes they can intersect.

Assessing individual instances in ABSA is further complicated by the predictions of multiple quadruples against several ground truth quadruples. 
This raises two salient questions: \textit{how should each instance be assessed when there's no perfect alignment of sequences? How should one measure the similarity score between predictions and ground truths?}
Furthermore, the evolution and ubiquitous adoption of generative models~\cite{radford2018improving,brown2020language,raffel2020exploring} introduces another layer of complexity. With these models, there's a greater potential for diverse answers. Thus, the pertinent question becomes: \textit{how should such diverse responses be evaluated? Should they be perceived and assessed under the same lens as the traditional extract-and-classify scenarios?}

To resolve the aforementioned complexities, this position paper discusses how diverse predictions and ground truths are evaluated and how we can evaluate the outputs of the GLM era.
In Section~\ref{sec:background}, we simply summarize the notation and the distinct subtasks inherent to the ABSA task. Note that we only deal with the essential components to discuss the evaluation-related topics in the context of this position paper. 
In Section~\ref{sec:paradigm}, 
we survey how ABSA has been conducted prior to the generative era and explore the ongoing trends, particularly emphasizing inferential attempts facilitated by pretrained and generative language models. 
Followingly, in Section~\ref{sec:evaluation}, we take a look at conventional ABSA evaluation schemes, including exact match and F1 metrics, while also shedding light on prevalent similarity measures used to evaluate term and expression correspondence. 
Lastly, in Section~\ref{sec:suggestion}, we compare evaluation schemes from various perspectives along with diverse examples and offer a suggestion on the future direction of ABSA evaluation. 
The primary goal of our manuscript is to spotlight challenges with the adoption of conventional ABSA evaluation in the emergent paradigm of generative models and 
to further establish a guidelight for assessing the model output with multiple span extraction problems.

\section{Background}
\label{sec:background}
\subsection{Four elements of ABSA}

Sentiment analysis is the study that analyzes individuals' sentiments and perceptions towards entities and their attributes~\cite{liu2012sentiment}. In the ABSA task, the opinion term ($o$) and sentiment polarity ($s$) belong to the prior sentiment-related elements, while the aspect term ($a$) and aspect category ($c$) go to the entity-related elements. The definition of these four elements can be described as follows:

\begin{itemize}
    \item \textbf{Aspect term ($a$)} refers to a target entity or object about which an opinion is expressed, usually represented by a word or phrase, but it can occasionally be implicit, as shown in example (B) of Fig.~\ref{fig:example}. 
    \item \textbf{Opinion term ($o$)} conveys a sentiment or viewpoint about the aforementioned aspect. This is usually described in words or phrases, but it can be implicit. 
    \item \textbf{Aspect category (\category{$c$})} classifies the aspect into a specific class selected from a predefined category set. 
    \item \textbf{Sentiment polarity (\sentiment{$s$})} is the sentiment the author has for the aspect, which is usually revealed by the opinion term $o$ and divided into three classes: \textit{positive, neutral}, and \textit{negative}.   
\end{itemize}

\subsection{Subtasks of ABSA}

\begin{table}[]
\centering
\resizebox{\columnwidth}{!}{%
\begin{tabular}{ll}
\toprule 
\textbf{Task} & \textbf{Output} \\ \midrule 
Aspect term extraction (ATE) & $a$ \\
Aspect-opinion pair extraction (AOPE) & $a, o$ \\
Aspect-sentiment pair extraction (ASPE) & $a$, \sentiment{$s$} \\
Aspect-sentiment triplet extraction (ASTE) & $a, o,$ \sentiment{$s$} \\
Aspect-category-sentiment detection (ACSD) & $a$, \category{$c$} \sentiment{$s$} \\
Aspect-sentiment quad prediction (ASQP) & $a$, \category{$c$}, $o$, \sentiment{$s$}
\\ \bottomrule
\end{tabular}%
}
\caption{Target output elements for ABSA tasks. The names of the tasks follow \citet{zhang2022survey}.}
\label{tab:subtasks}
\end{table}

As shown in Table~\ref{tab:subtasks}, several inference types can be identified when discussing ABSA, with the more prominent ones being: aspect term extraction (ATE)~\cite{hu-2004-A-dataset}, aspect-opinion pair extraction (AOPE)~\cite{fan-etal-2019-AO-towe-dataset}, aspect-sentiment pair extraction (ASPE), aspect-sentiment triplet extraction (ASTE)~\cite{peng2020-aste-dataset}, aspect-category-sentiment detection (ASCD), and aspect-sentiment quad prediction (ASQP). 
As a representative dataset, SemEval datasets~\cite{pontiki-etal-2014-semeval,pontiki-etal-2015-semeval,pontiki-etal-2016-semeval} introduces a wealth of resources applicable to ATE, ASPE, and ACSD tasks.
In a recent study, the ASQP~\cite{zhang-2021-asqp-paraphrase} task, which considers all four elements, has gained attention.
However, because all four factors should be extracted from a given sentence(s), there may be implicit cases where aspects or opinion terms do not appear explicitly in the input sentence(s). To encompass such implicit cases, datasets that additionally include ``\textit{NULL}'' spans, e.g., ACOS~\cite{cai-2021-acos} and MEMD-ABSA~\cite{cai-2023-memdabsa}, have been introduced. 
Note that datasets for the ASQP task can be utilized in all the subtasks mentioned earlier. In the scope of this position paper, we primarily mention the essential subtasks and benchmarks aligned with the ASQP task. Comprehensive surveys encapsulating a broader collection of ABSA models and benchmarks can be found in works by \citet{zhang2022survey} and \citet{chebolu2022survey}.

\section{ABSA Models in Paradigm Shift} 
\label{sec:paradigm}
\subsection{Inference De Facto}

Traditional inference process in ABSA artfully combines a variety of techniques~\cite{chebolu2022survey}. 
It blends named entity recognition (NER)-style span extraction, the linking of aspect and opinion terms, as depicted by the highlighted spans in Fig.~\ref{fig:example}, and the categorical decision-making process on both category and sentiment, as illustrated by the arrows in Fig.~\ref{fig:example}. 
Also, this process can either incorporate binary/ternary (sentiment) or multiple (category) labels in nature. Prior to the emergence of large-scale models, the tasks of extraction and prediction have been processed in distinctly separate stages. 
This bifurcation boosted methodologies akin to NER and categorization~\cite{saias-2015-sentiue,phan-ogunbona-2020-modelling}.

The advent and widespread utilization of BERT-style~\cite{devlin-etal-2019-bert} pretrained language models (PLMs) found their synergistic alignment with the aforementioned inferential framework. This harmony is exemplified in the extract-classify method, elucidated in the foundational ACOS paper~\cite{cai-2021-acos}. This approach merges span-tagging extraction with subsequent prediction layers —- all integrated within a single model.
Across these methodologies, there's a consistent theme: aspects and opinions are systematically extracted from the provided input, while categories and sentiments are predicted based on a predefined set of classes.

\subsection{Recent Attempts with GLM}

Employing generative models for ABSA inference has garnered significant attention, primarily driven by the advent of landmark GLMs like GPT-3~\cite{brown2020language} and sequence-to-sequence (Seq2seq) styled language generation techniques such as BART~\cite{lewis-etal-2020-bart} and T5~\cite{raffel2020exploring}.
Among them, T5 has gained paramount prominence. 
To bolster the accuracy of these generative models and curb tendencies for hallucination, researchers have begun experimenting with constrained decoding~\cite{de2020autoregressive} during the generation phase. This technique aids the model in selecting tokens directly from the input sentence or the category and sentiment sets. 
With the constrained decoding, T5-centric methodologies~\cite{zhang-2021-asqp-paraphrase,mao-2022-seq2path,bao-2022-ijcai-tree-gen,hu-etal-2022-dlo-template,gou-etal-2023-mvp} reign supreme in this domain. This typically involves feeding the model with instance-quadruple pairs for training and testing. The ultimate objective is to fine-tune T5 into a model adept at decoding the appropriate set of quadruples derived from a single instance.

The capabilities of GLMs have been advancing at a remarkable pace, and their applicability has broadened remarkably. 
Initially, GLMs found their primary niche in inherently generative tasks, such as machine translation, story generation, or dialogue management.
However, the unveiling of language modeling techniques that enable both comprehensive understanding and generative faculties, i.e., InstructGPT~\cite{ouyang2022training}, and its latecomers, brought about a novel approach such as in-context learning. 
They were also employed to proceed with categorical data; thus, metrics like accuracy and F1 score became instrumental in gauging their efficacy.

It is important to note that metrics for evaluating existing models have been performed separately for classification, e.g., precision, recall, and F1 scores, and generation, e.g., BLEU~\cite{papineni-etal-2002-bleu}, ROUGE~\cite{lin-2004-rouge}, and BERTScore~\cite{zhang2019bertscore}.
However, with the advent of GLMs, decoder-based models are actively used for classification tasks~\cite{min-etal-2022-noisy,min-etal-2022-rethinking,yoo-etal-2022-labelsdoesmatter-ICL}, in addition to generation tasks where decoder-only or encoder-decoder models have traditionally been used. 
In ABSA, the transition of the backbone model from BERT to T5 also echoes this trend.
Using decoder-based models for tasks where encoder-based models were traditionally applied has shifted the paradigm to performing classification and extraction from the generated outputs. 
This tends to blur the boundaries between conventional evaluation schemes.
Therefore, it is necessary to discuss how the evaluation method should be reflected concerning the change in the model paradigm.

\section{Evaluation Schemes}
\label{sec:evaluation}
\subsection{ABSA Evaluation Schemes}

As with the development of ABSA inference approaches, the evaluation landscape of ABSA has also been rich with distinct metrics tailored to the specific components of the analysis. For the categorical attributes, namely category and sentiment, widely accepted metrics such as accuracy and F1 score are employed. This decision is reasonable, given that the pool of candidates for these attributes is finite. However, it's worth noting that there exists an inherent disparity in the number of candidates across these attributes; while categories generally offer a broader selection compared to the sentiments, which are typically capped at three, albeit with the added layer of subjectivity.

When it comes to evaluating aspects and opinions, the exact match metric is the go-to~\cite{cai-2021-acos,gou-etal-2023-mvp}.
However, this metric is notably harsh for span outputs, primarily owing to the decision schemes associated with aspects and opinions, especially from the viewpoint of manual data construction. 
These schemes are often contingent on the domain of the corpus and the type of sentences.
Complicating matters further is the intrinsic challenge of distinguishing between aspects and opinions, and it is difficult to establish rules of thumb for consistently dissecting them in certain circumstances. For instance, if we have three sentences: \medskip\\
(1a) ``The \aspect{\textit{dinner}} was so \opinion{\textit{expensive}}.''\smallskip\\ 
(1b) ``The \aspect{\textit{price}} was \opinion{\textit{high}} in the dinner time.''\smallskip\\ 
(1c) ``The \aspect{\textit{dinner price}} was quite \opinion{\textit{high}}.'' \medskip\\
all the categories indicate the ``\category{Price}'' without doubt, but deciding the aspect and opinion for (1c) would be somewhat obscure given (1a) and (1b), which may invoke inconsistency in the annotation. Also, ``\textit{expensive}'' itself is the term that implies the information on pricing, which adds a challenge. If these kinds of discrepancies and challenges occur simultaneously in training and test sets, it would result in an inadvertent disadvantage of potentially appropriate predictions if the exact match is the only metric utilized.

For those seeking a more lenient evaluation metric for aspects and opinions, partial match~\cite{ku-etal-2008-question,li-etal-2022-multispanqa} schemes such as word-level F1 score can be an attractive alternative. 
Specifying the score into word or token level can provide a more delicate assessment than the exact match, which demands perfection.
It allocates scores even when there's a slight variance in expression, accommodating those instances that the exact match metric would deem incorrect.
On the other hand, given that the correct answer can be extracted in diverse forms, e.g., whether the GTs contain adverbs or not, we can refer to the dataset construction and evaluation methodology of machine reading comprehension (MRC) tasks.
In the case of SQUAD 2.0~\cite{rajpurkar-etal-2018-know-squad2} and KLUE-MRC~\cite{park-et-al-2021-klue}, the MRC task is performed to find the domain of a question in a given context, and several candidates are prepared according to the inclusion of articles such as `the', and the model prediction is considered successful if the output generated by the model exists in the candidates.
In ABSA task, it can be another alternative to consider the additional GT candidates for aspect and opinion terms. 

One crucial part of ABSA inference is that it capacitates multiple quadruple predictions and multiple ground truth answers (GTs) for a single instance. Presently, precision and recall are mainly investigated in the evaluation; that is, the evaluation process incorporates counting \textit{effective} predictions and all GTs that are correctly predicted. 
However, the evaluation doesn't necessarily factor in scenarios where the predictions do not necessarily exact match with GT but are informative enough to be assessed as an effective prediction. In such circumstances, both the exact match and partial metrics can impose a rather rigorous evaluation criterion.

\subsection{Studies on NLG Evaluation}

With the ascendancy of generative models as the standard for various downstream tasks, the shift towards incorporating natural language generation (NLG) evaluation schemes in lieu of traditional methodologies is both palpable and pragmatic. 
A substantial segment gravitates towards similarity metrics~\cite{sai2022survey}. However, while there exists a plethora of sentence-level metrics, such as BLEU~\cite{papineni-etal-2002-bleu} or ROUGE~\cite{lin-2004-rouge}, these don't seamlessly align with the task at hand, since neither the predictions nor the GTs are not necessarily sentence-level expressions.
Particularly at the level of entity or phrase, the word-level F1 score stands out in the representative area of question answering~\cite{rajpurkar-etal-2016-squad}, furnishing both precision and recall metrics for generated outputs.

Another avenue worth exploring is PLM-based approaches, such as BERTScore~\cite{zhang2019bertscore}. However, a caveat accompanies this strategy: it's profoundly sensitive to domain-specific influences, which might skew evaluations based on the domain's inherent characteristics, that may have discrepancy with the property of the pretraing corpus. 

Concerning the above deliberations, the consensus tilts towards adopting a variety phrase-level semantic similarity metrics that can be applied to aspect and opinion. Here, the partial match schemes discussed above can be again considered as a frontrunner. Its strength lies in calculating precision and recall based on the overall similarity of expression, which can compensate for the harshness of the exact match, and sometimes cover the cases when there is no overlap but the meaning is shared.

\section{Discussion}
\label{sec:suggestion}
\subsection{Comparison of Evaluation Schemes}

Considering all evaluation schemes that apply to current ABSA literature, we can come up with four following main topics of discussion.

\subsubsection{Using Exact Match vs. Partial Match} 
Transitioning into a generative paradigm introduces challenges in ensuring that generated terms align precisely with the intended GT spans. 
While strategies like constrained decoding can help an accurate exact match, it necessitates a reflection on the choice of evaluation metrics: should one adhere to the strict exact match criterion or explore more lenient alternatives like word-level F1 score or other similarity measures? 
For instance, in  (1c), given that the answer is ``\textit{dinner price}'', though exact match maximizes the utility of precise prediction of span extraction, devaluating predictions such as ``\textit{dinner}'' or ``\textit{price}'' without consideration on the approximation of semantics might be harsh concerning the potential utility of the model. 

Additionally, a subtle difference exists between various partial match metrics, which have pros and cons depending on domain, sentence/answer type, and dissecting schemes, e.g., whether to adopt whitespace, morphological decomposition, or other tokenization methodologies. Also, it is important whether to take into account the order of words in evaluation or not, which may lead to the superiority of metrics such as longest common substring (LCS)~\cite{li-etal-2022-multispanqa}.

\subsubsection{Total vs. Element-Wise Evaluation} 
When assessing quadruple predictions with GTs, the word-level F1 score doesn't account for the variances across individual elements. 
Given the distinct objectives and evaluative natures of different attributes, the element-wise evaluation might offer a more comprehensive assessment of model performance than the total one. 
This approach ensures that each attribute's unique characteristics and challenges are taken into consideration during the evaluation. 
For instance, in the same generative paradigm, evaluation on category and sentiment should better be an exact match considering that the predefined set of candidates is provided, while a partial answer should be tolerated for aspect and opinion terms given that the set of candidate sequences varies from input sentences. 

In sum, the quadruple-level assessment can shed light on the \textit{total} performance of the model, given that the score is added only if the whole inference is correct for all elements. 
However, it is questionable that aspect/category and opinion/sentiment are evaluated with the same criteria just because they are yielded as an output of a generative model.

\subsubsection{NLG Metrics} 
Concerning previous claims on the advantage of partial metrics, employing NLG evaluation metrics for span evaluation appears logical. However, delving deeper into them reveals inherent limitations. Metrics such as BLEU, ROUGE, and BERTScore are predominantly designed to evaluate sentences rather than isolated phrases.
This would be a useful metric for a special case if GT spans a whole sentence as an aspect or opinion term.

\subsubsection{Case Study}
\label{sec:casestudy}
\begin{table}[]
\centering
\resizebox{\columnwidth}{!}{%
\begin{tabular}{l|ll}
 & \textbf{Exact match} &\textbf{Partial match} \\ \toprule
\textbf{Total} & $ s \in \{0, 1\}$ & $ s \in {[}0, 1{]} $ \\ \midrule
\begin{tabular}[c]{@{}l@{}}\textbf{Element} \\ \textbf{-wise} \end{tabular}& 
\begin{tabular}[c]{@{}l@{}}$ s = \{s_i\}_{i=1}^{N_e} $ \\ $where\text{ } s_i \in \{0, 1\}$ \end{tabular} & 
\begin{tabular}[c]{@{}l@{}}$s = \{s_i\}_{i=1}^{N_e} $ \\ $where\text{ } s_i \in {[}0, 1{]}$ \end{tabular}
\end{tabular}%
}
\caption{Score formulation for four evaluation schemes. $s$ is the score, $N_e$ is the number of elements to be considered, where $N_e$ is 4 for ASQP task that $\{1,2,3,4\}$ corresponds with ($a$, \category{$c$}, $o$, \sentiment{$s$}).}
\label{tab:metric-comparison}
\end{table}

To compare the evaluation metrics with a detailed example, we sample a data instance from ACOS-laptop16 dataset.
Assume that we have a generative model, e.g., T5, and the model returns two quadruples for a single input sentence. The input, GT, and generated quadruples can be described as follows, where the extracted quadruples follow the order of ($a$, \category{$c$}, $o$, \sentiment{$s$}).\medskip\\
$x$ = ``\aspect{\textit{key}} presses are too \opinion{\textit{stiff}} to press .''\smallskip\\
$g$ = (\textit{key}, \category{Keyboard usability}, \textit{stiff}, \sentiment{Negative})\smallskip\\
$p_1$ = (\textit{key}, \category{Keyboard usability}, \textit{too stiff}, \sentiment{Negative})\smallskip\\
$p_2$ = (\textit{key presses}, \category{Keyboard usability}, \textit{stiff}, \sentiment{Negative})\smallskip\\
$x$ is an input sentence, $g$ is a GT quadruple of $x$, and $p_1$ and $p_2$ are the predictions of a model. 
Note that $p_1$ and $p_2$ have slightly different $o$ and $a$ mentions compared to $g$, respectively.

Now, let $f_n(g,p_i)$ be a score function of $g$ and $p_i$ that corresponds to four cases of Table~\ref{tab:metric-comparison}. The scoring metric is accuracy here.
Also, as shown in Table~\ref{tab:metric-comparison}, the total scores are scalar-valued, while the element-wise scores can be represented as sequences of length $4$ in this case study.

For total and exact match case (function $f_1$), the score of $p_1$ and $p_2$ are both zero, i.e., ${f_1}(g, p_i) = 0$, because the aspect and opinion terms are not exactly the same with the GT ($g$).
For total and partial match case (function $f_2$), the score of $p_1$ and $p_2$ are both $5/6$, i.e., ${f_2}(g, p_i) = 0.83$; note that for partial match scores, we count the common (whitespace-split) words between the GT and prediction.
While the total-exact match score concentrates on the wrongly predicted span, the total-partial match score shows the overall accuracy of the generated prediction. 

On the other hand, the element-wise scores in this ASQP case can be represented as a quadruple following the same order of ($a$, $c$, $o$, $s$).
The element-wise exact match case (function $f_3$) of $p_1$ is $(1., 1., 0., 1.)$ and that of $p_2$ is $(0., 1., 1., 1.)$, highlighting the element where the model wrongly generated.
However, these scores also reveal the harshness as a metric of the exact match.
Lastly, the element-wise partial match case (function $f_4$) of $p_1$ is $(1., 1., 0.5, 1.)$ and that of $p_2$ is $(0.5, 1., 1., 1.)$, showing the potential of predictions that would have been underestimated concerning other metrics.

To add complexity to the utility of partial matching metrics, the opinion terms with negating expressions (e.g., \textit{``no'', ``not'', ``less''}) can be considered as a challenging real-world example not handled in this case study. 
Imagine that a generative model (probably without constrained decoding) generates another prediction:\medskip\\
$p_3$ = (\textit{key}, \category{Keyboard usability}, \textit{not stiff}, \sentiment{Neutral}) \medskip\\
by inserting ``\textit{not}'' in the opinion term. 
In this case, the score of element-wise partial match($f_4$)  is $(1., 1., 0.5, 0.)$. 
However, one may find it questionable that the opinion score of $p_3$, namely $f_4(g,p_3)_3$, is the same as that of $p_1$, namely $f_4(g,p_1)_3$, which displays the opinion term semantically similar to the GT.
In this case, NLG metrics that consider semantic similarity would be an auxiliary measure that can penalize and filter out the predictions that provide contrary meanings and distort the evaluation. Simply in this example where the meaning of the opinion term of $p_3$ significantly contradicts that of GT, an NLG metric can assign 0, i.e., $f_4(g,p_3)=(1., 1., 0., 0.)$, for the sake of reasonable score assignment.

\subsection{Future Suggestion for Generative Paradigm}

Along with the four perspectives discussed above, we provide our viewpoints on each topic.

\paragraph{Partial match metrics should be supportively used.} 
The exact match metric, while fitting in pre-generative paradigms where span tagging and categorization were considered subsequent processes, might not be the sole reliable yardstick in a generative setting. Introducing metrics like word-level F1, LCS~\cite{li-etal-2022-multispanqa}, or edit distance could serve as indicators of the model's partial success, tempering the strict nature of the exact match. Ideally, a combination of these metrics would provide a more rounded perspective, showcasing both the precision and potential of predictions. We can also apply this idea to the element-wise evaluation that is to be discussed followingly.

\paragraph{Quadruple-level aggregation can be comprehensive, but element-wise evaluation can highlight the characteristics of each element.} 
Recognizing the distinct natures of each attribute is essential in evaluations. Especially the contrasts between categorical and span-based inferences demand acknowledgment. 
Even if the generative paradigm renders categorical outputs as textual representations rather than logits, their evaluation would be more suited to a categorical framework. One potential approach might involve separately evaluating the categorical attributes, i.e., category and sentiment, and the span-based ones, i.e., aspect and opinion. 
However, this does not basically touch the inherent variances in the challenges associated with each element. Typically, attributes like category and aspect present formidable hurdles in an accurate retrieval. This calls for different applications of similarity measures to each element: e.g., exact match for category/sentiment and word-level F1 for aspect/opinion. This can be considered a recommendable combination of various exact and partial metrics.

\paragraph{Exact quadruple match should accommodate partial match metrics in assessing prediction-GT pairs.} 
Evaluating entire quadruples collectively might obscure the nuances and variations that exist between individual predictions or GTs. Precision and recall, in this context, should be viewed through the lens of the prediction-GT similarity within individual quadruple pairs rather than searching for the existence of exactly the same output. Specifically, we can assume an adjusted scoring scheme based on a holistic comparison of entire predictions with their corresponding GTs and vice versa. For instance, if three predictions correspond closely to just one out of two GT quadruples, scores pertaining to the three prediction-GT pairs could be consolidated. Simultaneously, there should be a deduction in the aggregate score to account for the overlooked gold quadruple. As an example, we can think of the following concept of a system: \medskip\\
Let $P = \{p_1,p_2,p_3,p_4\}$  be a set of predictions for a single instance whose ground truth is $G = \{g_1,g_2,g_3\}$. In de facto exact match-based evaluation, precision and recall would be calculated concerning if $p_i \in G$ for all $i$ and if $g_j \in P$ for all $j$. 
However, assuming that $\{p_1,p_2\}$ were exact match or close guess to $g_1$ (in terms of a similarity metric), $\{p_3,p_4\}$ to $g_2$, and none of them were relevant to $g_3$, as following quadruples, which follow the order of ($a$, \category{$c$}, $o$, \sentiment{$s$}):\medskip\\
$p_1$ = (\textit{dinner}, \category{Price}, \textit{so expensive}, \sentiment{Negative})\smallskip\\ 
$p_2$ = (\textit{dinner price}, \category{Price}, \textit{so high}, \sentiment{Negative})\smallskip\\ 
$p_3$ = (\textit{beverage}, \category{Food quality}, \textit{too cold}, \sentiment{Negative})\smallskip\\ 
$p_4$ = (\textit{beverage}, \category{Drink}, \textit{cold}, \sentiment{Positive})
\medskip\\
$g_1$ = (\textit{dinner}, \category{Price}, \textit{so expensive}, \sentiment{Negative})\smallskip\\ 
$g_2$ = (\textit{beverage}, \category{Drink}, \textit{too cold}, \sentiment{Negative})\smallskip\\ 
$g_3$ = (\textit{lamb steak}, \category{Food quality}, \textit{awesome}, \sentiment{Positive})\medskip\\
set aside from the harsh exact match metric which may give the precision of 0.25 and the recall of 0.33, we can think of weight averaging all relevant similarity scores regarding $(p_1,g_1), (p_2,g_1)$ and $(p_3,g_2), (p_4,g_2)$ as a correspondence to $\{g_1,g_2\}$ and penalize the whole score for not even getting close to $g_3$. Note that this concept is just a recommendation; though it does not strictly follow the conventional exact match scheme, it accommodates partial exact match metrics and also allows an element-wise analysis.

\paragraph{Using NLG metrics might be necessary, but needs to be considerate.}
Current ABSA evaluations predominantly steer clear of NLG metrics tailored for sentence-level similarities, such as BLEU, ROUGE, or BERTScore. Given that most GTs and predictions exist at the word or phrase level, the fit might seem misaligned. However, as we discussed in Section~\ref{sec:casestudy}, there are scenarios where these metrics can be relevant, considering span extraction's alignment with extractive summarization. Similarly, if the GT for an opinion is ``\textit{quite small}'' and the prediction is ``\textit{not big enough}'', the degree of correctness becomes debatable. If the span extraction or constrained decoding mechanisms are employed, such discrepancies might be less probable to arise, but in their absence, should this variance diminish? Additionally, these metrics could offer insights into the implications of minor word deviations in incorrect predictions. Without a doubt, these metrics require a deep consultation that depends on model types and element properties.

\section{Conclusion}

In this paper, we delved deep into the complexities of aspect-based sentiment analysis (ABSA) and its evaluative mechanisms, especially in the evolving landscape of generative models. Drawing from existing literature, we explored the intricacies of ABSA inference methodologies and existing evaluation schemes, highlighting their strengths and limitations. Through our discussion, we underscored the inherent challenges of aligning generative outputs with stringent evaluative metrics, emphasizing the need for a more delicate approach that factors in the generative paradigm's unique attributes. While we don't prescribe a singular metric, our exploration offers insights into the benefits and potential pitfalls of various methodologies. Ultimately, our paper aims to serve as a compass, offering guiding directions for practitioners navigating the intricate terrains of ABSA inferences within the generative paradigm.

\section*{Limitation}
\label{sec:limitation}

Our focus in this discussion has been predominantly on refining the metrics for ACOS predictions, particularly considering the inclusion of the \textit{NULL} entity as proposed in the original paper. However, it's essential to acknowledge the limitations of our approach. In this discourse, we've exclusively explored cases pertaining to explicit aspects and opinions, deliberately sidelining instances with implicit terms. Such cases, rich in their inherent complexities, are earmarked for future exploration and analysis.

Additionally, it's worth noting that our intention here isn't to prescribe a definitive metric. Rather than pinpointing an optimal direction for a specific objective, our endeavor has been to shed light on the advantages and disadvantages of various methodologies. Our aim remains to offer a balanced perspective, equipping practitioners with insights that can guide their evaluative processes.

\bibliography{custom}

\appendix



\end{document}